\documentclass{article}
\usepackage{arxiv}

\usepackage[utf8]{inputenc} 
\usepackage[T1]{fontenc}    
\usepackage{hyperref}       
\usepackage{url}            
\usepackage{booktabs}       
\usepackage{amsfonts}       
\usepackage{nicefrac}       
\usepackage{microtype}      
\usepackage{lipsum}         
\usepackage{graphicx}
\usepackage{natbib}
\usepackage{doi}
\usepackage{amsmath}
\usepackage{cleveref}       
\usepackage{todonotes}

\newcommand{\Ni}{({\it i\/})~}
\newcommand{\Nii}{({\it ii\/})~}
\newcommand{\Niii}{({\it iii\/})~}

\newcommand{\real}{\ensuremath{\mathbf{R} }}

\title{Learning Physical Concepts in CPS: A Case Study with a Three-Tank System} 

\usepackage{booktabs}				
\usepackage{makecell}

\usepackage{wasysym}  
\usepackage{url}  
\author{
    \href{https://orcid.org/0000-0002-7812-4279}{\includegraphics[scale=0.06]{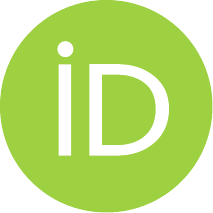}\hspace{1mm}Henrik S.~Steude} \\
	Institute of Automation Technology\\
    Helmut Schmidt University\\
    Holstenhofweg 85, 22043 Hamburg\\
	\texttt{henrik.steude@hsu-hh.de} \\
	\And
\href{https://orcid.org/0000-0002-6522-4262}{\includegraphics[scale=0.06]{orcid.pdf}\hspace{1mm}Alexander Windmann} \\
	Institute of Automation Technology\\
    Helmut Schmidt University\\
    Holstenhofweg 85, 22043 Hamburg\\
	\texttt{alexander.windmann@hsu-hh.de} \\
	\And
\href{https://orcid.org/0000-0002-6522-4262}{\includegraphics[scale=0.06]{orcid.pdf}\hspace{1mm}Oliver Niggemann} \\
	Institute of Automation Technology\\
    Helmut Schmidt University\\
    Holstenhofweg 85, 22043 Hamburg\\
	\texttt{oliver.niggemann@hsu-hh.de} \\
}

\hypersetup{
pdftitle={Learning Physical Concepts in CPS: A Case Study with a Three-Tank System},
pdfauthor={Henrik S.~Steude, Alexander Windmann, Oliver Niggemann},
pdfkeywords={Machine Learning, Representation Learning, Physical Concept Identification}
}

\begin{document}
\maketitle

\begin{abstract}                
Machine Learning (ML) has been implemented with great successes in recent decades, both in research and in practice.
In the field of Cyber-Physical Systems (CPS), ML methods are already widely used, e.g. in anomaly detection, predictive maintenance or diagnosis use cases.
However, Representation Learning (RepL), which learns general concepts from data and has been a major driver of improvements, is hardly utilized so far.
To be useful for CPS, RepL methods would have to \Ni produce interpretable results \Nii work without requiring much prior knowledge, which simplifies the implementation and \Niii be applicable to typical CPS datasets, including e.g. noisy sensor signals and discrete system state changes.
In this paper, we provide an overview of the current state of research regarding methods for learning physical concepts in time series data, which is the primary form of sensor data of CPS.
We also analyze the most important methods from the current state of the art using the example of a three-tank system.
Based on concrete implementations\footnotemark, we discuss the advantages and disadvantages of the methods and show for which purpose and under which conditions they can be used for CPS.
\end{abstract}


\section{Introduction}
\label{sec:Intro}

\footnotetext{Code available at:\\ \url{https://github.com/hsteude/Concept-Learning-for-CPS}}

The success of Deep Learning \citep{Bengio2021-os} is enabled by the ability of neural networks to learn abstract concepts from complex data.
Once learned, these abstract concepts can simplify the solution of a wide variety of problems. Often, these concepts correspond to unobservable or even unknown quantities. Famous examples are BERT \citep{Devlin2018-dm} for natural language processing problems and ResNet \citep{He2016-zt} for computer vision tasks.
However, concept learning and transfer learning are hardly used for CPS.

In most cases, CPS data are available in the form of multivariate time series.
If deep learning models were available that could extract or identify physical concepts from this sensor data, simpler solutions could be found for tasks such as predictive maintenance \citep{Nguyen2019-dx}, anomaly detection \citep{Niggemann2015-dy} or diagnosis \citep{Zhang2019-do}.
Examples of such physical concepts would answer the following questions:
 Can the large number of sensors be represented by a small number of variables that describe the state of the system?
 If so, how can these variables be interpreted?
And can the behavior of the CPS over time be described in a simple way?
These questions motivate this paper and lead to the following research questions: 
\Ni Can the benefit of RepL methods be demonstrated on CPS examples? What are useful concepts for CPS models? How are concepts defined for time series?
\Nii What are the advantages and disadvantages of these methods and to which CPS use cases can they be applied?

\begin{figure}[htb!]
    \centering
    \includegraphics[scale=0.8]{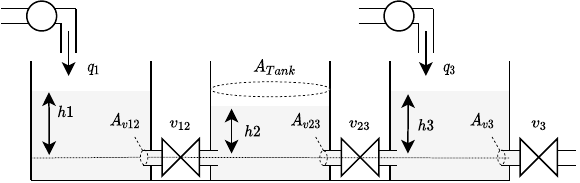}{\caption{
        This figure schematically shows the three-tank system, which we will use as a reference system throughout this paper.
        It includes the unobservable inflow $q_1(t)$ and $q_3(t)$, the observable variables $h_1(t), h_2(t), h_3(t)$, the valves $v_{12}, v_{23}, v_3$, the cross-sectional areas of the individual pipes and the connections between the tanks.
        These quantities are examples of concepts that learning algorithms should ideally identify from the multidimensional time series of sensor data of the system.
        The current system state or a description of the temporal behavior of this system are also examples of potentially learnable concepts.
        }}\label{fig:three-tank-overview}
\end{figure}

To answer these questions, we implement four concept learning methods that we consider promising using the three-tank system as an example (see Figure \ref{fig:three-tank-overview}).
This example is well suited in our context for two reasons:
First, the complexity is low and the concepts are therefore relatively easy to understand.
Second, it can be easily extended incrementally in complexity up to the Tennessee Eastman Process \citep{Balzereit2021-vy}.

The remainder of this paper is organized as follows:
In Section \ref{sec:SOTA}, we provide an overview of the current state of research related to solving research question (\textit{i}).
Based on this, in Section \ref{sec:Solution} we present and discuss our implementations.
In order to discuss the methods' advantages and disadvantages, in Section \ref{sec:Results} we evaluate the methods with respect to their applicability to typical CPS datasets, the interpretability of the respective results, and the amount of prior knowledge required.
In addition, we consider the application of the methods for typical CPS use cases.
Finally, Section \ref{sec:Conclusion} concludes this paper.

\section{State of the Art}
\label{sec:SOTA}

The task of learning physical concepts using ML methods can be seen as a subset of the very active and relatively new research area of RepL.
The core motivation of RepL is to build models which are capable of encoding real world observations of (physical) processes into meaningful representations \citep{Bengio2013-yv}.
Good representations both increase transparency and simplify downstream tasks such as predictions or classifications.
Often, this means that a multivariate time series $\mathbf{X} \in \mathbf{R}^{s\times n}$, where $s$ denotes the sequence length and $n$ the number of sensors, is mapped to a vector $\mathbf{z}\in\mathbf{R}^{m}$ whith $m \ll n$,  such that  $\mathbf{z}$ encodes the disentangled explanatory factors of the observation $\mathbf{X}$.
In most cases, computing representations is accompanied by a dimensionality reduction.
The idea of dimensionality reduction is already very old.
Methods such as PCA \citep{Hotelling1933-qe} and MDS \citep{Kruskal1964-tt} have been used successfully for decades.
New methods for dimensionality reduction, such as Stochastic Neighbor Embedding \citep{Hinton2002-xx}, t-SNE \citep{Van_der_Maaten2008-zj}, and UMAP \citep{McInnes2018-pm} are also frequently applied in the field of CPS.

However, these methods seldom encode meaningful representations and even more rarely physical concepts.
For this purpose, strong priors are required.
In modern RepL, several of these priors are used and studied, see \citep{Bengio2013-yv}.
A common prior is to assume simple and sparse dependencies between a small number of underlying factors that explain the variation in the observed data.
Probably the best known example is the assumption that the underlying explanatory factors are statistically independent.
This prior lies at the core of deep generative models such as Variational Autoencoders (VAEs) \citep{Kingma2013-js} and Generative Adversarial Networks (GANs) \citep{Goodfellow2014-qq}. 
VAEs and GANs have been supplemented and improved in many ways over the past few years \citep{Higgins2017-aw,Kim2018-iz,Chen2018-zh,Chen2016-yx,Zhao2019-yd}.
While corresponding methods have achieved impressive success in RepL for high-dimensional data such as images or videos, stronger priors are required for identifying physical concepts, especially in complex CPS datasets.

Within the field of RepL, there has been a number of works attempting to learn discrete representations. Popular examples include restricted Boltzmann machines \citep{RuslanSalakhutdinov.2009} and VQ-VAE \citep{vandenOord.2017}.
Discrete representations are a good fit for CPS, as the system's behavior can often be described by underlying states. Furthermore, discrete representations are more interpretable and thus simplify supervision. By applying these methods on time series, the dynamic behavior of the underlying system can be analyzed, as demonstrated by SOM-VAE \citep{fortuin2018deep}. Successful applications on CPS include the use of self-organizing maps for predictive maintenance \citep{Birgelen.2018} and restricted Boltzmann machines for anomaly detection \citep{Hranisavljevic2020-nw}.

\cite{Nautrup2020-af} as well as \cite{Iten2020-ji} present a newer algorithm specifically designed for the identification of physical concepts.
The authors present a method that contains a number of decoder and encoder neural networks (called agents) that exchange information among each other.
While the encoding agents map the observations into a representation, the decoding agents perform different subtasks using the representations as input.
These subtasks are chosen such that each decoding agents requires a different subset of underlying explanatory factors to perform their task.
By means of a special loss function, the communication between the agents is minimized, which disentangles the individual variables in the latent space.

The literature also provides ML approaches that aim at extracting physical concepts in the sense of simple and sparse symbolic formulas.
An important example is Symbolic Regression (SR), where the aim is to find a symbolic expression that maps some given input to some output data.
A very popular solution is described in  \citep{Schmidt2009-xf} and implemented in the Eureqa software. 
More recently, a new method called AI Feynman was introduced \citep{Udrescu2020-ka, Udrescu2020-tb}.
A use case for SR, which is very promising especially in the context of CPS, is the discovery of dynamic systems.
E.g. \cite{Brunton2016-ic} propose a method called SINDy, which uses SR methods to discover parsimonious models for non-linear dynamic systems.

An important assumption of the SR based methods mentioned above, however, is that the sparse symbolic expression can be found in the coordinate system in which the observations are measured.
A new field of ML research is emerging around the question of how suitable coordinate system and dynamic systems can be learned simultaneously.
\cite{Champion2019-ly} e.g. use an autoencoder architecture in combination with the SINDy algorithm. 
The autoencoder maps the observations from some high-dimensional space into a coordinate system that enables a parsimonious representation of the system dynamics in the latent space.
Because linear dymanic systems are of great advantage over non-linear systems for control engineering and prediction, there are also approaches that allow the identification of linear dymanic models in latent space \citep{Lusch2018-jw}.
This approach is very closely related to the Koopman operator theory.
A recent and comprehensive literature review on this topic is provided by \cite{Brunton2021-uf}.

To the best of our knowledge, there is no paper systematically evaluating the application of current deep RepL methods in the field of CPS.

\section{Solution}
\label{sec:Solution}
In this section, we analyze the application of a selection of the methods mentioned in Section \ref{sec:SOTA} in the field of CPS.
For this purpose, we consider the example of the three-tank system (see Figure \ref{fig:three-tank-overview}).
The dynamics of this system with respect to the fill levels of the tanks can be described as follows \citep{Kubalcik2016-vy}:
\begin{equation}
\label{eq:dynamics}
\begin{aligned}
    \dot{h}_1 &= \frac{1}{C}q_1 - \frac{k_{v_{12}}}{C}\text{sign}(h_1 - h_2)\sqrt{|h_1- h_2|}\\
    \dot{h}_2 &= \frac{k_{v_{12}}}{C}\text{sign}(h_1 - h_2) \sqrt{|h_1- h_2|}\\
              &\quad- \frac{k_{v_{23}}}{C}\text{sign}(h_2 - h_3) \sqrt{|h_2- h_3|}\\
    \dot{h}_3 &= \frac{1}{C}q_3 - \frac{k_{v_{23}}}{C}\text{sign}(h_2 - h_3)\sqrt{|h_2 - h_3|} - k_{v_3}\sqrt{h_3}\\
\end{aligned}
\end{equation}
where $h_1, h_2$ and $h_3$ are the time dependent fill levels of the corresponding tanks, $q_1$ and $q_3$ are the flow rates of the pumps, $k_{v_{12}}, k_{v_{23}}$ and $k_{v_3}$ are the coefficients of the valves $v_{12}, v_{23}$ and $v_{3}$ respectively and $C$ is some system specific constant.
Ideally, the methods would identify explanatory but unobservable physical quantities such as the inflow or valve coefficients based on observations of this system.
Ideally, even the system dynamics or process phases like ``mixing'' or ``filling'' could be identified.

\subsection{Solution 1: Seq2Seq variational autoencoder}
\label{subsec:VAE}
Deep generative models such as VAEs and GANs can be trained to extract the underlying factors of variation in data.
These models often approximate the joint probability $p(\mathbf{X}, \mathbf{z})$, where $\mathbf{X} \in \mathbf{R}^{s\times n}$ represents the observations and $\mathbf{z} \in \mathbf{R}^{m}$ the latent space variables, which encode the explanatory factors of variation in $\mathbf{X}$.
The core assumption in this approach is that the underlying factors of variation in the data also describe the underlying physical concepts causing the observations.
In this case, the physical concepts would be encoded in the form of latent variables $\mathbf{z}$.

\subsubsection{Experiment}
\begin{figure}[htb!]
    \centering
    \includegraphics[scale=.65]{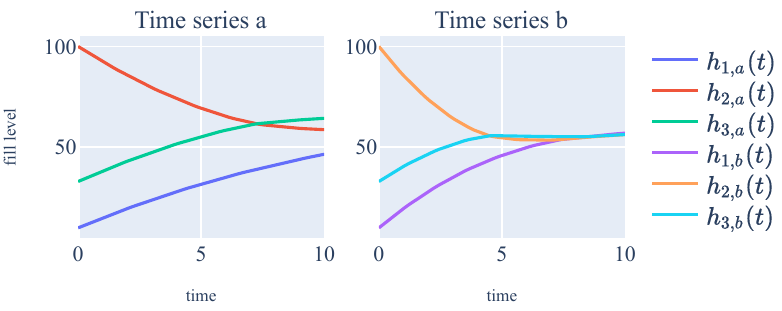}{\caption{
            Example time series from the dataset used to train the VAE. 
            The values for time series a and b are set to $q_1 = 1, q_3 = 2, k_{v_{12}}= 1, k_{v_{23}} = 0.5$ and $q_1 = 1, q_3 = 2, k_{v_{12}}= 1, k_{v_{23}} = 0.5$ respectively.
            For both time series the initial condition of the system at $t=0$ is $x_0 = [h_{1,0}, h_{2,0}, h_{3,0}]^T = [30, 10, 90]^T$.
    }\label{fig:vae-ts-plot}}
\end{figure}

\begin{figure}[htb!]
    \centering
    \includegraphics[scale=.7]{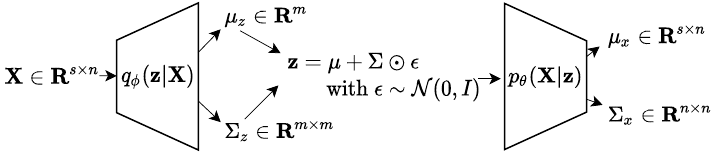}{\caption{
            Architecture of the Seq2Seq VAE.
            The input $\mathbf{X}$ is a $n$-dimensional time series with length $s$.
            The output is a Gaussian probability distribution $p_\theta(\mathbf{X|z})$.
            During training we optimize the parameters $\theta$ and $\phi$ such that the likelihood of the training samples $\mathbf{X}_i$ under $p_\theta(\mathbf{X}_i|\mathbf{z}_i)$ is maximized and the KL-Divergence between $q_\phi(\mathbf{z}_i|\mathbf{X}_i)$ and the prior $\mathbf{z}_i\sim \mathcal{N}(0, 1)$ is minimized at the same time.
    }\label{fig:vae-architecture}}
\end{figure}

To demonstrate this method, we simulate a dataset based on the dynamics described in Equation \eqref{eq:dynamics}.
The dataset contains $10000$ different time series of lenght $s=50$, each describing an independent process of the three-tank system.
The individual time series differ in the values of the inflows $q_1$ and $q_3$ as well as in the valve coefficients $k_{v_{12}}$ and $k_{v_{23}}$.
We sample these quantities randomly from a given interval. Figure \ref{fig:vae-ts-plot} shows two example time series from this dataset.
\\
Using this dataset, we train a $\beta$-VAE \citep{Higgins2017-aw}.
We have adapted the solution from the original paper to better handle time series data by using Gated Recurrent Units (GRU) \citep{Cho2014-eu} in the encoder $q_{\phi}(\mathbf{z}|\mathbf{X})$ and decoder $p_{\theta}(\mathbf{X}|\mathbf{z})$ with parameters $\phi$ and $\theta$ respectively (see Figure \ref{fig:vae-architecture}). 
Given the continuous nature of the concepts we are interested in, we choose a Gaussian prior such that $\mathbf{z} \sim \mathcal{N}(0, I_5)$ and $\text{dim}(\mathbf{z}) = 5$.

\subsubsection{Results}
\begin{figure}[htb!]
    \centering
    \includegraphics[scale=.65]{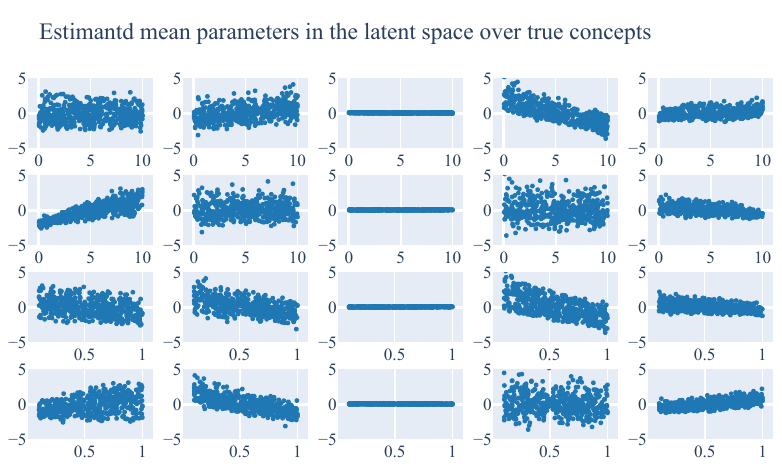}{\caption{
            This scatter plot matrix shows the correlations between all possible pairs of the estimated mean $\mu_z$ of the latent space distributions $p_\theta(\mathbf{z}| \mathbf{X})$ and the actual physical quantities $q_1, q_3, k_{v_{12}}$ and $k_{v_{23}}$.
            The rows correspond from top to bottom to the underlying physical concepts ($q_1, q_3, k_{v_{12}}$ and $k_{v_{23}}$) and the columns from left to right to the activations of the neurons in the latent space ($\mu_{z_1}, \mu_{z_2}, \mu_{z_3}, \mu_{z_4}, \mu_{z_5}$).
    }\label{fig:vae-results-plot}}
\end{figure}

Evaluating the quality of representations is not trivial in general.
In our example, however, we are able to compare the mappings learned by the model with the actual physical factors underlying the data.
In the case of good representations in our sense, a strong correlation should be observed between one of the estimated mean parameters $\mu_z$ of the distributions $p_\theta(\mathbf{z}| \mathbf{X})$ and one of the true underlying quantities $q_1, q_3, k_{v_{12}}$ and $k_{v_{23}}$.
In other words, each of the physical concepts should be reflected in the activation of one of the latent neurons.
At the same time, this activation should be as indifferent as possible to changes in the other physical quantities.
This can be seen to some extent in Figure \ref{fig:vae-results-plot}.
We can observe that one of the five latent variables encodes almost no information from the input.
This makes sense, since it is only four independent variables that cause the changes in the data.
A clear correlation can be seen in three of the subplots, showing the scatter plots of the pairs $(q_1, \mu_{z_4})$, $(q_{3}, \mu_{z_1})$ and $(k_{v_{32}}, \mu_{z_2})$.
However, no clear disentanglement of the individual concepts emerges, although the corresponding conditional likelihood $p_\theta(\mathbf{X}|\mathbf{z})$ is very high.

\subsubsection{Discussion}
As this experiment shows, deep generative models can be trained in a purely data-driven manner to learn representations of time series data.
The fact that the prior knowledge required for training is limited to the choice of the prior distribution and the number of latent neurons is a clear advantage of this method.
Another advantage is that this method can also be applied to high-dimensional data.
As this experiment shows, deep generative models can be trained in a purely data-driven manner to learn representations of time series data.
The fact that the prior knowledge required for training is limited to the choice of the a priori distribution and the number of latent neurons is a clear advantage of this method.
Another advantage is that this method can also be applied to high-dimensional time series data, which are often available in the CPS environment.
A key disadvantage of this approach, however, is that it is not specifically designed to identify physical concepts, but rather to identify those factors that explain the variation in the observations.
Furthermore, it is a disadvantage, that the concepts learned are limited to physical quantities. Concepts such as event sequences or formulas cannot be identified.

\subsection{Solution 2: Communicating agents}
\label{subsec:CommAgents}
\begin{figure}[htb!]
    \centering
    \includegraphics[scale=.7]{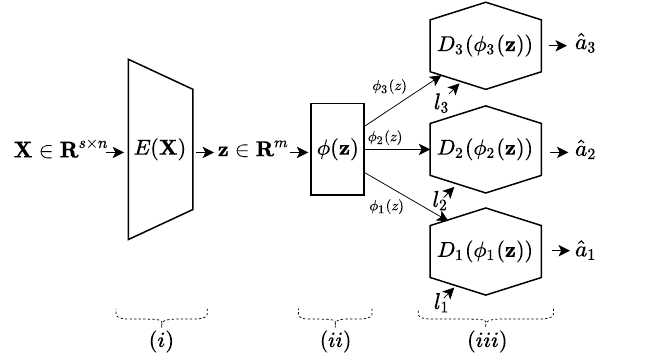}{\caption{
            High level overview over network architecture of the communicating agents setup including the encoder (\textit{i}), the filter \Nii and the decoding agents $(iii)$.
    }\label{fig:comm-agents-architecture}}
\end{figure}

In most cases strong priors are needed to extract actual physical concepts from CPS data.
In this section we demonstrate a method introduced by \cite{Nautrup2020-af}. 
Their solution essentially consists of three components (see Figure \ref{fig:comm-agents-architecture}).
\Ni There is (at least) one encoder neural network $E(\mathbf{X})$ that maps the observations $\mathbf{X} \in \real^{s \times n}$ into a lower dimensional space $\mathbf{z} \in \real^m$ using a deep neural network.
\Nii There are $k$ decoders $D_1(\mathbf{z}), \dots, D_k(\mathbf{z})$ that solve different tasks (questions) using the latent representation $\mathbf{z}$ of the observations as input.
\Niii A filter component $\phi(\mathbf{z})$, which limits the amount of information transferred from the encoder to each individual decoder. 
Intuitively, the filter can be thought of as adding random noise to the latent space variables before transferring them to the decoders.
The loss function maximizes the amount of noise added, while minimizing the mean squared error of the regression problems solved by the decoders.

\subsubsection{Experiment}
To generate data for this experiment, we use the same simulation as in Subsection \ref{subsec:VAE}.
However, training this method also requires the dataset to contain ``questions'' $l_1, \dots l_k$ and ``answers'' $a_1, \dots, a_k$, which are essentially labels.
Training in this constellation only helps to disentangle the latent variables if different physical concepts are required to answer the different questions.
For our experiment, we assume that it is possible to answer the following questions for each training sample:
 Given some flow rate $q_1$ (or $q_3$), what is the time it takes to fill up Tank 1 (or Tank 3) when all the valves are closed ($l_1$ and $l_2$)?
And if either Tank 1 (or Tank 2) is completely filled and all valves are opened, how long will it take for the system to drain ($l_3$ and $l_4$)?
To generate the complete dataset, we compute the answers $a_1, \dots, a_4$ to these four questions in advance.
In contrast to the authors of the original paper, we use GRUs in the encoder neural network.

\subsubsection{Results}
\begin{figure}[htb!]
    \centering
    \includegraphics[scale=.65]{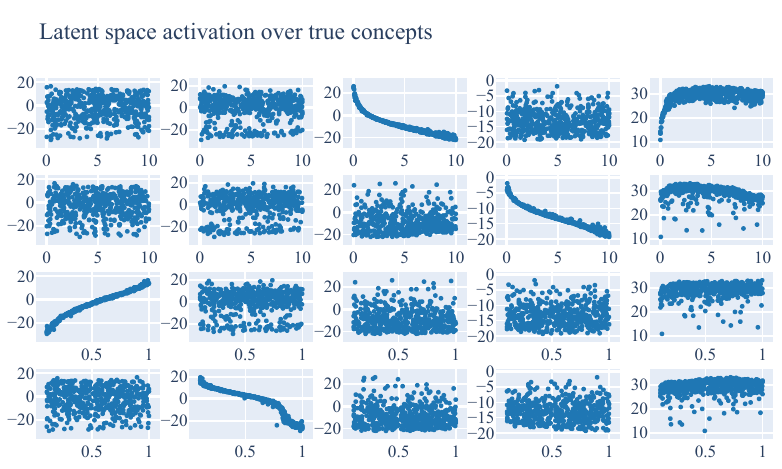}{\caption{
            This scatter plot matrix shows the correlations between all possible pairs of the latent space activation $\mathbf{z}$ and the actual physical quantities $q_1, q_3, k_{v_{12}}$ and $k_{v_{23}}$.
            The rows correspond from top to bottom to the underlying physical concepts ($q_1, q_3, k_{v_{12}}$, $k_{v_{23}}$) and the columns from left to right to the activations of the neurons in the latent space ($z_1, z_2, z_3, z_4, z_5$).
        }\label{fig:comm-agents-results}}
\end{figure}

In Figure \ref{fig:comm-agents-results}, we see that the correlations between the activations of the latent space neurons and the physical concepts are much stronger in this experiment than in the previous one.
This result was to be expected, because with the dataset enriched with questions and answers, a stronger prior was set to disentangling the latent variables.

\subsubsection{Discussion}
The experiment shows that the method produces a better disentanglement between the individual concepts underlying the data than the VAE.
Thus, the possibility of incorporating prior knowledge through the design of experiments has advantages.
However, the need for datasets that include questions and answers is also the major drawback of this approach, as performing experiments in CPS is rarely feasible.
In addition, as with the VAE, only physical quantities can be identified, but not their interaction or temporal sequences.

\subsection{Solution 3: Dynamic system identification}
\label{subsec:SINDy}
\begin{figure}[htb!]
    \centering
    \includegraphics[scale=.9]{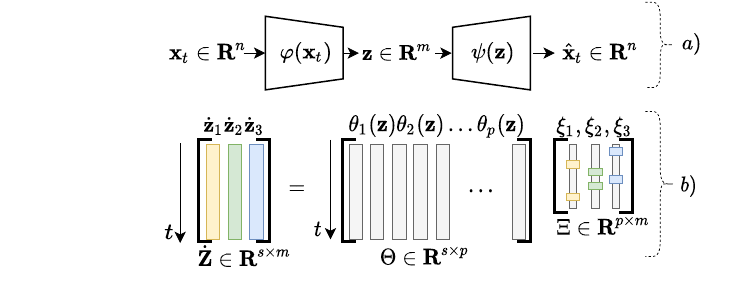}{\caption{
            Model architecture of the Autoencoder-SINDy algorithm.
            $a)$ Shows the autoencoder setup.
            In this part of the figure, only one point in time is considered.
            $b)$ Illustrates the SINDy algorithm in the latent space. 
            In this part, the whole time series is considered.
        }\label{fig:autoenc-sindy-architecture}}
\end{figure}

In this section, we will use the SINDy and the Autoencoder-SINDy method to show how the system dynamics of our example system can be identified.
For both methods the derivatives $\dot{\mathbf{X}}$ must be available or computed numerically prior to the model training.
The SINDy algorithm basically performs a sparse regression to approximate $\dot{\mathbf{x}}_t=f(t,\mathbf{x_t})$, where $\mathbf{x}_t \in \mathbf{R}^{n}$ is one snapshot in time and one row of $\mathbf{X}\in \mathbf{R}^{s \times n}$.
The method allows the user to define a library of $p$ possible candidate functions $\theta_1, \dots, \theta_p$, which are used to create a feature matrix $\Theta(\mathbf{X})\in \real^{s\times p}$.
This matrix is used to identify sparse parameters of the coefficient matrix $\Xi\in \real^{p\times m}$ such that $\Theta(\mathbf{X}) \Xi \approx f(\mathbf{X}) = \dot{\mathbf{X}}$ (see Figure \ref{fig:autoenc-sindy-architecture} $b$). 
The SINDy method assumes that a dynamical model can be identified in the observed variables.
The Autoencoder-SINDy algorithm on the other hand assumes that there is a non-linear coordinate transformation $\varphi(\mathbf{x}_t)=\mathbf{z}$ that allows the formulation of the system dynamics as a simple mathematical expression $f(\mathbf{z}) = \dot{\mathbf{z}} \approx \Theta(\varphi(\mathbf{x}_t))\Xi$ in the latent space.
Equation \eqref{eq:sindy-loss} shows the loss function $\mathcal{L}_{AS}$ of the Autoencoder-SINDy setup.
$\lambda_1,  \lambda_2$ and $\lambda_3$ are the weights for the SINDy-loss in $\dot{\mathbf{x}}_t$, SINDy-loss in $\dot{\mathbf{z}}$ and the regularization penalty respectively.
\begin{equation}
\label{eq:sindy-loss}
\begin{aligned}
    \mathcal{L}_{AS} &= \underbrace{\|\mathbf{x}_t - \psi(\mathbf{z})\|_2^2}_{\text{reconstruction loss}}
                \hspace{-3pt}+\hspace{3pt} \lambda_1\underbrace{ \| \mathbf{\dot{x}}_t - (\nabla_z\psi(\mathbf{z}))(\mathbf{\Theta}(\mathbf{z}^T)\mathbf{\Xi})\|_2^2}_{\text{SINDy loss in }\dot{\mathbf{x}}_t} \\
                &+ \lambda_2\underbrace{ \|\nabla_x\varphi(\mathbf{x}_t)\mathbf{\dot{x}}_t - \mathbf{\Theta}(\varphi(\mathbf{x}_t)^T)\mathbf{\Xi}\|_2^2}_{\text{SINDy loss in }\dot{\mathbf{z}}} \hspace{8pt}+\hspace{-14pt} \underbrace{\lambda_3\|\mathbf{\Xi}\|_1}_{\text{SINDy regularization}}
\end{aligned}
\end{equation}
During training, the weights of $\varphi$ and $\psi$ as well as the coefficient matrix $\Xi$ are optimized with respect to the loss function above.

\subsubsection{Experiment}
For this example, we also use the dynamical system (Equation \eqref{eq:dynamics}) to simulate training data.
We assume that all valves except Valve 3 are open and that the system is in some random initial state $\mathbf{x}_0 = [h_{1,0}, h_{2,0}, h_{3,0}]^T$.
To generate the complete dataset we sample 1000 different initial conditions and run the simulation for 50 time steps and obtain a total dataset of 50000 training samples. 
For the SINDy experiment, we assume that the levels of the tanks can be observed.
For the Autoencoder-SINDy experiment on the other hand, we assume that we observe the simulation in a higher dimensional space.
To generate this high-dimensional observation, we use a polynomial transformation of degree 5 such that $\mathbf{x} = \left[1, z_1, z_2, z_3, z_1^2, z_1z_2,z_1z_3, z_2^2, \dots, z_3^5\right]$.

\subsubsection{Results}
For the SINDy experiment we used the Python implementation available on GitHub \citep{De_Silva2020-hf}.
As expected, the quality of the results strongly depends on the choice of candidate functions.
If we include $\theta(a,b) = \text{sign}(a-b)\sqrt{|a-b|}$  in the  library in addition to the typical candidates such as polynomials and trigonometric functions, we can easily identify Equation \eqref{eq:dynamics}.
However, if this rather specific function is not part of the candidate library, no sparse expression can be identified.

\begin{figure}[htb!]
    \centering
    \includegraphics[scale=0.6]{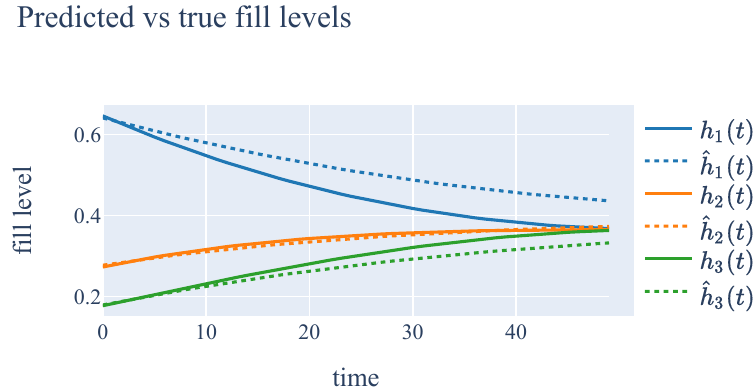}{\caption{
        This plot shows the results of the integration of the ODE identified by the Autoencoder-SINDY model and the actual time series of observations of the system.
    }\label{fig:ae-sindy-results}}
\end{figure}

The Autoencoder-SINDy model has identified the following equation for the dynamics in the latent space:
\begin{equation}
\begin{aligned}
    \dot{z_1} &= -0.051z_0 + 0.003z_1 - 0.014 \text{sin}(z_0) \\
              & \quad+ 0.003 \text{cos}(z_0) + 0.007z_0z_1  \\
    \dot{z_2} &=  0 \\
    \dot{z_3} &= -0.008z_0 - 0.043z_2 + 0.008 z_1z_2 - 0.004z_2^3\\
              &\quad -0.001 \text{sign}(z_2) \sqrt{z_2} \\
\end{aligned}
\label{eq:AE-SINDy-result}
\end{equation}
This equation differs significantly from Equation \eqref{eq:dynamics}.
However, it is worth noting that only two latent variables have a nonzero change over time.
This makes sense because e.g. the differences of the fill levels $h_1-h_2$ and $h_2-h_3$ would suffice to describe the dynamics of the system in our experiment.
To validate the quality of the model, we conduct the following experiment:
First, we take some initial value $\mathbf{x}_0$ and compute $\hat{\mathbf{z}_{}}=\varphi(\mathbf{x}_{0})$.
We then solve the ODE in Equation \eqref{eq:AE-SINDy-result} to get a time series $\hat{\mathbf{Z}}\in\mathbf{R}^{50\times3}$.
Finally we transform each value in the resulting time series back to the observation space using $\hat{\mathbf{X}}=\psi(\hat{\mathbf{Z}})$.
As Figure \ref{fig:ae-sindy-results} shows, the resulting time series is very similar to the original sample. 
\subsubsection{Discussion}
This experiment shows that it is possible to extract simple formulas for dynamical systems from the observations in the case of both observable and unobservable system states.
In particular, the result of the SINDy experiment, namely the correct underlying differential system equation, increases the interpretability and scalability of predictive models compared to standard neural network approaches.
The selection of the candidate functions allows the integration of prior knowledge into the model.
However, both methods are very sensitive to the choice of candidate functions and the hyperparameter settings.


\subsection{Solution 4: State identification}
\label{subsec:state}

The state encodes the system's current behavior in a low-dimensional representation. Ideally, this representation has a topologically interpretable structure, i.e. states that are close are more similar. Such a topological structure can be induced by self-oganizing maps (SOMs)\citep{Kohonen.1990}.
A popular time series clustering method that builds on this approach is SOM-VAE \citep{fortuin2018deep}.
The model first uses an encoder to generate a representation  $\mathbf{z}_e \in \mathbf{R}^{m}$ from an observation $\mathbf{X} \in \mathbf{R}^{s\times n}$. In the latent space, the encoding is assigned to its nearest embedding vector $\mathbf{z}_q \in \mathbf{R}^{m}$. The embedding vectors are randomly initialized and have a pre-defined number and relative position. These embedding vectors function as a representation of the underlying state. At its core, the model is an autoencoder, thus a decoder uses both the encoding and the embedding vector to reconstruct the input, resulting in $\mathbf{\hat{X}_e}\in \mathbf{R}^{s\times n}$ and $\mathbf{\hat{X}_q}\in \mathbf{R}^{s\times n}$ respectively. While training, the model adjusts both encoding and embedding vectors in order to minimize the reconstruction loss, which is the first term of the loss function, see Equation \eqref{eq:somvae}. 
Furthermore, the encoding should be similar to its assigned embedding vector, which is handled by the commitment loss, the second term in the loss function.
Lastly, the topological structure induced by the SOM has to be learned. In essence, the neighbors $N\left(\mathbf{z}_q\right)$ of the assigned embedding vector $\mathbf{z}_q$ are pulled towards the encoding of the input. Crucially, the encoding $\mathbf{z}_e$ does not receive information about these other embeddings, which is noted by the gradient stopping operator $\operatorname{sg}[\cdot]$:
\begin{equation}
\label{eq:somvae}
\begin{aligned}
\mathcal{L}_{\text {SOM-VAE }}
&=\underbrace{\left\|\mathbf{{X}}-\mathbf{\hat{X}_q}\right\|^{2}+\left\|\mathbf{{X}}-\mathbf{\hat{X}_e}\right\|^{2}}_{\text{reconstruction loss}} +\hspace{4pt}\alpha\hspace{-4pt}\underbrace{\left\|\mathbf{z}_e-\mathbf{z}_q\right\|^{2} }_{\text{commitment loss}} \\
&\quad+\beta \underbrace{\sum_{\mathbf{\tilde{z}} \in N\left(\mathbf{z}_q\right)}\left\|\mathbf{\tilde{z}}-\operatorname{sg}\left[\mathbf{z}_e\right]\right\|^{2} }_{\text{SOM loss}}
\end{aligned}
\end{equation}

\subsubsection{Experiment}
To simulate changing states in the three-tank system described in Equation \eqref{eq:dynamics}, the values of the flow rates and the valve coefficients are changed periodically in a fixed sequence. 
The process can be seen in Figure \ref{fig:state-detection}, where the tanks are filled and mixed in stages, until the whole fluid is released. In total there are four different states: the tanks are filled, the flow is stopped with closed valves, the fluids of the tanks are mixed and finally all valves are opened, which causes the tanks to empty. The first three states are repeated three times in a row, followed by the last state.
This sequence is repeated for 100 times, resulting in a total of 145000 time steps. To generate the training dataset, 10000 samples with a window size of 100 are sampled from the dataset randomly.
After training the model, we iterate over the test dataset to generate the predicted state at every time step. To ensure that the model does not use future values, at every step the model only has access to the past 100 time steps, thus simulating a live prediction. 
Using this dataset, we train a SOM-VAE with the encoder and decoder implemented as fully connected dense neural networks. The model is given the possibility to assign a total of six different states, which are ordered in a $2\times 3$ grid. 
We deliberately do not choose a $2\times2$ grid here because the user generally does not know the underlying number of states.

\subsubsection{Results}
While the underlying settings of the flow rate and the valve coefficients are the same in every respective state, the fill level of the tanks accumulate, which means that the same state can have different fill levels. 
Ideally, a model should be able to detect the underlying state regardless of the differences in the fill levels. 

\begin{figure}[htb!]
	\centering
	\includegraphics[scale=.6]{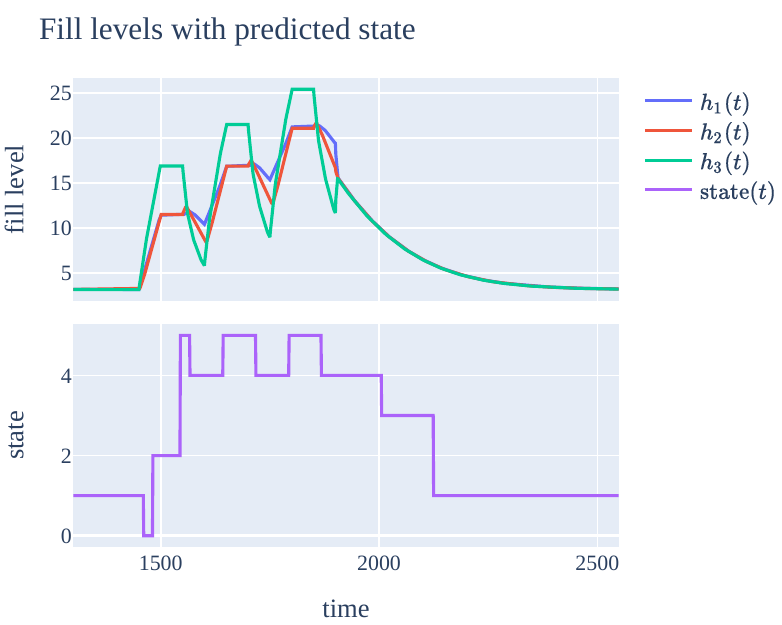}{\caption{
            Section of the dataset simulating a dynamically changing system. The top graph shows the process of filling and mixing the three-tank system in stages. The bottom graph shows the predicted states learned by SOM-VAE.
		}\label{fig:state-detection}}
\end{figure}

The predicted states at each time step can be seen in Figure \ref{fig:state-detection}. The first thing to notice is that the predictions have a small time lag. That is expected, as the model only receives information about past values and needs some time to adapt to changes.
The model detects the cycling through the filling, resting and mixing phases and switches between State 4 and 5 repeatedly. It can thus to some extend learn to generalize the system's behavior, as the individual fill levels differ in between each phase. However, the model struggles to differentiate between the filling and the mixing phase, which might occur too quickly. The model was given the task to assign six states, while the underlying system only has four. Some states thus do not seem to encode relevant information and pass quickly. 

\subsubsection{Discussion}
The model is able to detect changing states of the system in a fully unsupervised matter. Not only does it assign a sample to a state, it learns an embedding that represents the system's current behavior and that could be used in downstream tasks. Furthermore, the transition of states can be analyzed to gain insights in the temporal structure of the system, as demonstrated in the original SOM-VAE paper. 
While the underlying change of states has been detected, the predictions have a small time lag and quick transitions have not been detected properly.
Furthermore, the number of possible states an individual model can learn is a hyperparameter which needs to be set beforehand.


\section{Discussion} 
\label{sec:Results}
This section discusses the usefulness of concept learning for CPS.
We compare the methods demonstrated above on the basis of the following three core criteria:
\Ni \textit{applicability on CPS data}, which describes the ability of the methods to handle typical CPS data, including e.g. noisy measurements, discrete system state changes, and hybrid data forms, 
\Nii \textit{required prior knowledge}, which describes the amount of prior knowledge needed to apply the method, 
and \Niii \textit{interpretability}, the degree to which the results of the methods are interpretable.

The comparison is summarized in Table \ref{tab:comparison_method}, where $\CIRCLE$ indicates a high performance of the solution (e.g. highly interpretable results or no prior knowledge needed).
$\LEFTcircle$ and $\Circle$ indicate medium and low performance respectively.
\begin{table}[ht!]
	\centering
	\caption{Solution comparison---method}
	\label{tab:comparison_method}
	\renewcommand*{\arraystretch}{1.5}
	\begin{tabular}{@{} l c c c c c @{}}
		\toprule
		
		\textbf{Criterion} & 
		\thead{Sol. 1} &
		\thead{Sol. 2} &
		\thead{Sol. 3} &
		\thead{Sol. 4} \\
		\midrule
		\textbf{CPS data} & \CIRCLE & \CIRCLE & \Circle & \CIRCLE\\
		\textbf{Prior knowledge} & \CIRCLE &  \Circle & \LEFTcircle & \LEFTcircle\\
		\textbf{Interpretability} & \LEFTcircle & \CIRCLE & \CIRCLE & \LEFTcircle\\
		\bottomrule
	\end{tabular}
\end{table}\\
Leaving aside the necessary prior knowledge, it can be said that Solutions 1, 2 and 4 can be applied to typical CPS datasets very well.
These solutions have in common that the encoder can theoretically extract the representations from any complex time series dataset.
In contrast, Solution 3 assumes a continuous dynamical system underlying the observations.
However, in most CPS datasets there will be discrete sensor signals and externally triggered discrete state changes. 
With regard to the prior knowledge needed, Solution 2 stands out, as it requires labeled datasets which are hard to collect in a real-life CPS.
Solutions 1, 3, and 4 only require prior knowledge in the form of hyperparameters such as the choice of candidate functions (Solution 3) or the number of states (Solution 4).
The representations offered by Solutions 2 and 3 are most interpretable, as they either focus on physically meaningful latent variables (Solution 2) or symbolic expressions (Solution 3).

Another dimension along which the solutions can be compared is the degree to which their results can be used for different CPS use cases (see Table \ref{tab:comparison_uc}).
For simplification, we generalize the multitude of CPS use cases into three areas: \Ni \textit{system monitoring}, which describes all tasks related to monitoring CPS e.g. anomaly detection, \Nii \textit{prognosis}, which also includes simulation and predictive maintenance and \Niii \textit{diagnosis}, including all applications enabling root cause analyses related to system anomalies and failures.
\begin{table}[ht!]
	\centering
	\caption{Solution comparison---use cases}
	\label{tab:comparison_uc}
	\renewcommand*{\arraystretch}{1.5}
	\begin{tabular}{@{} l c c c c c @{}}
		\toprule
		
		\textbf{Use case} & 
		\thead{Sol. 1} &
		\thead{Sol. 2} &
		\thead{Sol. 3} &
		\thead{Sol. 4} \\
		\midrule
		\textbf{System monitoring} & \CIRCLE & \LEFTcircle & \CIRCLE & \CIRCLE\\
		\textbf{Prognosis} & \LEFTcircle &  \LEFTcircle & \CIRCLE & \LEFTcircle\\
		\textbf{Diagnosis} & \Circle & \LEFTcircle & \LEFTcircle & \LEFTcircle\\
		\bottomrule
	\end{tabular}
\end{table}\\
All solutions have in common that they include a dimension reduction. 
Monitoring a few ideally meaningful variables to assess the overall system state is generally easier than visualizing a large number of sensor signals in the observation space.
In addition, all solutions, with the exception of Solution 2, contain an autoencoder, which can be used for anomaly detection.
With respect to prognosis-related use cases, the lower-dimensional representations of all solutions might be helpful when used as input features for downstream ML models.
This is especially the case when the downstream tasks require labels and only a subset of the available data is labeled.
Furthermore, the generative model of Solution 1 can be used for simulations by analyzing the effects of changes in the latent space on the overall system. 
Solution 3 stands out, however, because it can be used explicitly to predict the behavior of the system over time by means of the system dynamics equation.
Finally, none of the solutions is readily suitable for diagnostic use cases.
However, the information gained by applying the solutions might simplify the root cause analysis.

\section{Conclusion}
\label{sec:Conclusion}

In this paper we have investigated to what extent deep RepL methods can be used in the field of CPS. 
We identified four different solutions to learn concepts in the data.
Using a simple three-tank system as an example, we tested a selection of algorithms and discussed their advantages and disadvantages.
We showed that, for example, VAEs and communicating agents can be used to extract the most important physical quantities from multidimensional CPS sensor data.
In addition, we demonstrated how to identify discrete system states with a SOM-VAE and showed that the Autoencoder-SINDy method can identify a mathematical expression describing the system dynamics.
Thereafter, we discussed the significance of each method in terms of its utility and applicability in CPS.

By applying recent algorithms from RepL on CPS, we have been able to show shortcomings of the solutions.
An interesting direction for future research would be to combine the methods for a better fit to the characteristics of CPS data.
For example, learning a symbolic expression could be greatly enhanced if the latent variables encode interpretable physical quantities, as demonstrated by the communicating agents.
Additionally, by filtering for discrete state shifts, the complexity of the dynamic system can be greatly reduced. 
This paper has mainly focused on learning concepts with interpretable representations.
In contrast, huge ML models that are trained on large datasets learn useful (but not interpretable) representations, which can be used to transfer knowledge to subsequent ML models. Likewise, transferring knowledge of physical concepts could improve ML models on CPS data. 
We believe this paper has shown the potential of concept learning and can motivate the development of algorithms that focus on the unique challenges CPS pose.

\bibliographystyle{unsrtnat}
\bibliography{review-concept-learning}
\end{document}